\documentclass[11pt]{article}
\usepackage{ijcnlp2011}
\usepackage{times}
\usepackage{url}
\usepackage{latexsym}
\usepackage{booktabs}
\usepackage[pdftex]{graphicx}

\title{Applying Deep Belief Networks to Word Sense Disambiguation}

\author{Peratham Wiriyathammabhum \footnotemark \\
Department of Computer Engineering,\\
 Chulalongkorn University\\
  {\tt calzera.p@gmail.com}
\And
  Boonserm Kijsirikul\\
  Department of Computer Engineering,\\
 Chulalongkorn University\\
  {\tt  boonserm@cp.eng.chula.ac.th} 
\\
\AND
 Hiroya Takamura\\
 Precision and Intelligence Laboratory\\
Tokyo Institute of Technology\\
{\tt takamura@pi.titech.ac.jp} 
\And
Manabu Okumura\\
Precision and Intelligence Laboratory\\
Tokyo Institute of Technology\\
{\tt oku@pi.titech.ac.jp}
}
\begin{document}
\maketitle
\begin{abstract}
  In this paper, we applied a novel learning algorithm, namely, Deep Belief Networks (DBN) to word sense disambiguation (WSD). DBN is a probabilistic generative model composed of multiple layers of hidden units. DBN uses Restricted Boltzmann Machine (RBM) to greedily train layer by layer as a pre-training. Then, a separate fine tuning step is employed to improve the discriminative power. We compared DBN with various state-of-the-art supervised learning algorithms in WSD such as Support Vector Machine (SVM), Maximum Entropy model (MaxEnt), Na\"ive Bayes classifier (NB) and Kernel Principal Component Analysis (KPCA). We used all words in the given paragraph, surrounding context words and part-of-speech of surrounding words as our knowledge sources. We conducted our experiment on the SENSEVAL-2 data set. We observed that DBN outperformed all other learning algorithms.
\end{abstract}

\section{Introduction}

A major difficulty of Natural Language Processing is to automatically resolve many ambiguities arising in human language, for instance, lexical ambiguity. When we put a polyseme into a sentence in order to communicate with other people, it is difficult for human to specify the meaning of that polyseme especially when there are many polysemes in a document. For example, a word "snow leopard" can refer to either an animal or a Macintosh operating system. However, we can look at the surrounding words to guess the meaning. By this guessing, we can use machine to help us disambiguate those polysemes in a document. 

Word sense disambiguation (WSD) is a task to computationally identify the appropriate meaning {\em s} from the given set of meaning {\em S} for a word {\em w} in a given context {\em c}. WSD is considered to be a fundamental task to achieve a high performance in Machine Translation (MT). Other applications of WSD include Information Retrieval (IR), Information Extraction (IE) and text mining.

There are four approaches for WSD which are knowledge-based methods, unsupervised corpus-based methods, supervised corpus-based methods and combinations of those approaches. In this paper, we focus on the supervised corpus-based approach which has been constantly observed as the highest performance gainer. A supervised approach starts with building feature vectors then employing learning algorithms for those feature in a classification.   

\footnotetext{This work started when this author was at 
  Interdisciplinary Graduate School of Science and Engineering,
 Tokyo Institute of Technology.}

Feature vectors can be constructed from the text in which the word {\em  w} has occurred. To begin with, the correct senses of the word {\em  w} in each context will be manually tagged and used as a label. Then, knowledge sources will be considered to make a feature such as part-of-speech or local bigram. Consequently, we will get one feature vector for each context and will be used as a training set to train a classifier for each word {\em  w}. There is an official competition which is conducted once in three years. The data sets from this competition are SENSEVAL-1, SENSEVAL-2, SENSEVAL-3, SEMEVAL-1 and SEMEVAL-2\footnote{http://www.senseval.org/}. In this paper, we considered the English lexical sample task of SENSEVAL-2 which has 73 word tasks including tasks for nouns, verbs and adjectives. SENSEVAL-2 used WordNet 1.7 to label the data. There are 75 to 300 instances in each word task.

The goal of a learning algorithm is to predict an unseen example correctly using knowledge from previously seen examples. Until now, the learning algorithms that have been shown to work well in WSD are Na\"ive Bayes (NB), Nearest Neighbors (NN), and Support Vector Machine (SVM). However, those learning algorithms are all 'shallow' learning algorithms. Shallow learning algorithms mean that the learning algorithms do not consist of nonlinearity that is complex enough to model human behaviors. Shallow learning algorithms may be effective when used to create a simple system. For example, it may succeed in one problem with a lot of human works in feature engineering but this system will be task specific and could not be reused for a new problem even if the problem is similar to the previous one. In addition, the feature vectors, which are the input of shallow learning algorithms, are sparse and cause the curse of dimensionality problem.  

Deep learning algorithms aim at learning feature hierarchies where higher level features are formed by the composition of lower level features. Although the features are constructed in a recursive manner, each feature level represents a different level of abstraction. This is important for extraction of higher level abstractions where human cannot explicitly specify the system. Thus, deep learning may be used in addition to typical feature engineering for natural language processing where the system will have more coverage because a feature extractor from deep learning can be generalized to similar problems. Until now, there are many proposed deep learning algorithms; however, this paper will investigate the behavior of Deep Belief Networks (DBN).

In this paper, we conducted an experiment to compare various ‘shallow’ learning algorithms with DBN on basic features of the SENSEVAL-2 English lexical sample data set.

\section{Related Works}

Lee and Ng (2002) evaluated various learning algorithms with many knowledge sources and the result claimed that a linear Support Vector Machine (SVM) is the best classifier. Escudero (2006) also investigated the effectiveness of the Linear SVM. We et al. (2004) introduced Kernel Principal Component Analysis (KPCA) with polynomial kernels to find a nonlinear combination of features for classifiers. The result showed that Na\"ive Bayes (NB), Maximum Entropy (MaxEnt) and SVM all got better performance.

There are several works that apply neural networks to WSD. Cottrell (1989) was the first who proposes neural networks for WSD. However, Towell and Voorhees (1998) argued that neural networks without a hidden layer have better performance. This goes with the previous statement which concluded that WSD data is likely to be linear and sparse. Thus, Linear SVM would be the best classifier for WSD.

Recently, deep learning algorithms consistently showed interesting results over shallow algorithms in many natural language processing tasks. Collobert and Weston (2008) proposed a deep neural network architecture which can be applied to part-of-speech tagging, chunking, name entity recognition and semantic role labeling simultaneously. The proposed architecture learns internal representation and shares that representation as a feature among tasks. Mnih and Hinton (2008) proposed a deep neural network for language model which outperforms non-hierarchical neural models and n-gram language models. Other successes are machine transliteration (Deselaers et al., 2009), sentiment analysis (Zhou et al., 2010;Glorot et al., 2011), question answering (Wang et al., 2010), named entity recognition (Chen et al., 2010), relation extraction (Chen et al., 2010), Parsing (Socher et al., 2010). As far as we know, there is still no investigation with recently advanced deep learning in WSD.

In WSD, it is empirically shown that linear SVM works best so the structure of the data seems to be linear. However, this work will address the possibility that WSD data may be nonlinear and the performance can be improved when using deep learning algorithms even if the number of instances per class is small and the feature vector is highly sparse.

\section{Knowledge Sources}

\subsection{Topical Feature}
We collected all unigrams in the provided context whether they were in the different sentences or not and encoded them to a binary bag-of-words feature vector. We used the word segmentation module and Porter stemmer module from NLTK (Bird et al., 2009) for preprocessing. We also used stop words list from NLTK to remove stop words. This type of feature defines a general topic of the text which comes from an intuition that the words in the same topic usually occur together.

\subsection{Local Feature}
We specified the size of window which covers around the target word {\em w} needed to be disambiguated. The window will produce the words before and after the word {\em w}. The typical window size is between 3 to 10 words. This feature type encoded the position of words in local vicinity. We included local unigram, bigram and trigram in order to construct the feature vector. For example, from the phrase "cross the river", we will have three unigrams (cross, the, river), two bigrams (cross\_the, the\_river) and one trigram (cross\_the\_river). We used the word segmentation module from NLTK (Bird et al., 2009) for preprocessing. Finally, we got the binary feature which represented the local feature. In the experiment, we found that the window size of 7 yielded the best performance.

\subsection{Part-of-speech Feature}
We used the part-of-speech tagger module from NLTK (Bird et al., 2009) to tag all unigrams in the specified window. Then, we encoded them to binary features which represented the position of the part-of-speech tag of each word. For instance, assume that we have four part-of-speech, NN (Noun), VB (Verb), ADJ (Adjective) and DT (Determiner). If we have the tagged phrase "cross/VB the/DT river/NN", the feature vector will be \begin{math} \langle 0,1,0,0,0,0,0,1,1,0,0,0 \rangle \end{math}. First four digits encode the part-of-speech of the word "cross" and so on.  In this feature, we used only the words in the same sentence as the target word {\em  w}.

\section{Learning Algorithms}

We evaluated following seven learning algorithms in order to compare with Deep Belief Networks. Those learning algorithms are Na\"ive Bayes, Nearest Neighbors, Principal Component Analysis, Kernel Principal Component Analysis, Logistic Regression (MaxEnt), Multilayer Perceptron and Support Vector Machine. In this section, we denote {\em  x} as a data instance and {\em  y} as a label instance. {\em  X} and {\em  Y} are matrics where each column is a data instance {\em  x} or label instance {\em  y} respectively.

\subsection{Na\"ive Bayes}
Na\"ive Bayes (NB) is a simple learning algorithm which illustrates the use of Bayes rule with the assumption that all features are conditionally independent given a class. NB chooses the class with highest posterior probability as a prediction. In the experiment, we used Naïve Bayes module from NLTK (Bird et al., 2009) and used Laplace (add-one) smoothing.

\subsection{Nearest Neighbor}
A nearest Neighbor classifier (NN) classifies by choosing the closest training example in the feature space. NN are often regarded as lazy learning since the computation will be done only when classification. A k-Nearest Neighbor algorithm takes a majority vote among its {\em k} neighbors. However, NN will be extremely slow of the data having many instances or many dimensions. In the experiment, we used Nearest Neighbor algorithm from scikits.learn\footnote{http://scikit-learn.sourceforge.net} and set the {\em  k} parameter to 1.

\subsection{Principal Component Analysis }
Principal Component Analysis (PCA) is a dimensionality reduction technique. PCA maps data points to the feature space while preserving as much variance as possible. PCA solves the eigenvalue problem of the zero-mean covariance matrix {\em  C} to find eigenvectors {\em  V} ordered by descending magnitude of the corresponding eigenvalues \begin{math}\lambda\end{math} and uses them as bases for projection.
\begin{equation}
C = cov(x) 
\end{equation}
\begin{equation}
CV = \lambda V
\end{equation}
 The target feature space usually has very small dimension compared to the original feature space. In the experiment, we implemented PCA by using NumPy and SciPy (Jones et al., 2001). We specified target dimension to 30 and used 1-NN as a classifier.

\subsection{Kernel Principal Component Analysis }
Kernel Principal Component Analysis (KPCA) extends PCA to nonlinearity. KPCA introduces kernel trick where the data is mapped to reproduce kernel Hilbert space (RKHS) which is a convenient way to model nonlinearity by implicit mapping. KPCA computes kernel matrix {\em  K} using the kernel function \begin{math} \kappa \end{math}.
\begin{equation}
K = \kappa (x_i, x_j)
\end{equation}
Matrix K is double-centered by the following equation,
\begin{equation}
K_{ij} = K_{ij} - \frac{1}{n} \sum_{p} K_{ip} -  
\frac{1}{n} \sum_{q} K_{qj} + \frac{1}{n^2} \sum_{pq} K_{pq}
\end{equation}
Then, KPCA solves the eigenvalue problem like PCA.
\begin{equation}
KV = \lambda V
\end{equation}
The bases for projection are the eigenvectors scaled by the square root of their corresponding eigenvalues.
\begin{equation}
\alpha_i = \frac{1}{\sqrt{\lambda}} v_i
\end{equation}
Noted that, the test data are needed to be double-centered with the training data before projection.

In the experiment, we implemented KPCA by using NumPy and SciPy (Jones et al., 2001). We specified target dimension to 30 and used 1-NN as a classifier. We experimented with Gaussian RBF and polynomial kernels.

\subsection{Logistic Regression}
Logistic Regression applies the technique of linear regression to the classification problem in a probabilistic way. Logistic Regression could be considered as an instance of Maximum Entropy model (MaxEnt). The objective function of Logistic Regression is to minimize the prediction error of the prediction :
\begin{equation}
y_{predict} = argmax_i P(Y=i|x,W,b),
\end{equation}
where {\em  W} is the weight matrix and {\em  b} is the bias. Probability for each class is the value of the softmax function of the input.
\begin{equation}
P(Y=i|x,W,b) = \frac{e^{Wx_i+b}}{\sum_{i}e^{Wx+b}}
\end{equation}

In the experiment, we employed Logistic Regression from Theano (Bergstra et al., 2010) which used Stochastic Gradient Descent (SGD) to optimize the loss function. We fixed the learning rate of 0.13 for SGD.

\subsection{Multilayer Perceptron}
Multilayer Perceptron (MLP) is a feedforward artificial neural network model. MLP consists of input layer, hidden layer and output layer. Feature vector will be viewed as an input layer where each feature corresponds to an input node. The hidden layer tries to transform the input feature vector by learning. The output layer takes the output of hidden layer as an input and acts as a classifier. MLP uses backpropagation algorithm (BP) for learning. BP adjusts weights with respect to the gradient of an error measure. The error in the output unit is computed first, and the error is propagated through all layers. In the experiment, we used MLP with one hidden layer from Theano (Bergstra et al., 2010) and used Stochastic Gradient Descent (SGD) to optimize the loss function. We fixed the learning rate of 0.01 for SGD. We used one hidden layer with 1,000 nodes.
\begin{figure}[h]
\begin{center}
\includegraphics[scale = 0.32]{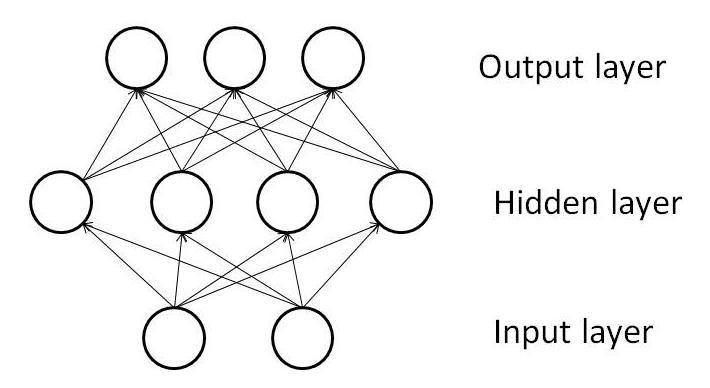}
\caption{\label{font-figure} The architecture of Multilayer Perceptron.}
\end{center}
\end{figure}

\subsection{Support Vector Machine}
Support Vector Machine (SVM) finds an optimal separating hyperplane for two class classification where the margin is widened as possible by using quadratic programming. If two classes are nonseparable, parameter {\em  C} will be used for controlling the tradeoff between the width of the margin and training error as follows: 
\begin{eqnarray}
\min_{W,b,\varepsilon_i} \lambda {\| W \|}^2 + C \sum_{i=1}^n \varepsilon_i  \qquad \quad \\
s.t. \; y_i (\langle W, x_i \rangle+b) \ge 1-\varepsilon_i , \, and \, \varepsilon_i > 0 \, \forall i.
\end{eqnarray}
SVM may be extended by the kernel trick to support nonlinearity in the data set in the same sense as KPCA but in the dual forms of SVM optimization problem. 

In the experiment, we used SVM from scikits.learn which made a function call to LIBSVM (Chang and Lin, 2011) and LIBLINEAR (Fan et al, 2008). We set the parameter {\em  C} to 1. For nonlinear SVM, we used third order polynomial kernel and Gaussian RBF kernel with parameter \begin{math} \Gamma \end{math} (gamma) set to  3.

\section{Deep Belief Networks}
Deep Belief Networks (DBN) (Hinton, 2006) are graphical models which extract hierarchical representation from the data. DBN consists of multiple layers of binary stochastic latent variables. The learning steps of DBN start by greedily learning the feature layer by layer one layer at a time using a kind of Markov Random Field (MRF) called Restricted Boltzmann Machine (RBM). The learned hidden layer will be used as an input layer for another layer recursively. The objective of this phase is to find a good parameter set for DBN which is used as an initial parameter for the second phase which all layers will be fine-tuned with the backpropagation algorithm (BP) to improve discriminative power. The second step will adjust all parameters in all layers.
\begin{figure}[h]
\begin{center}
\includegraphics[scale = 0.32]{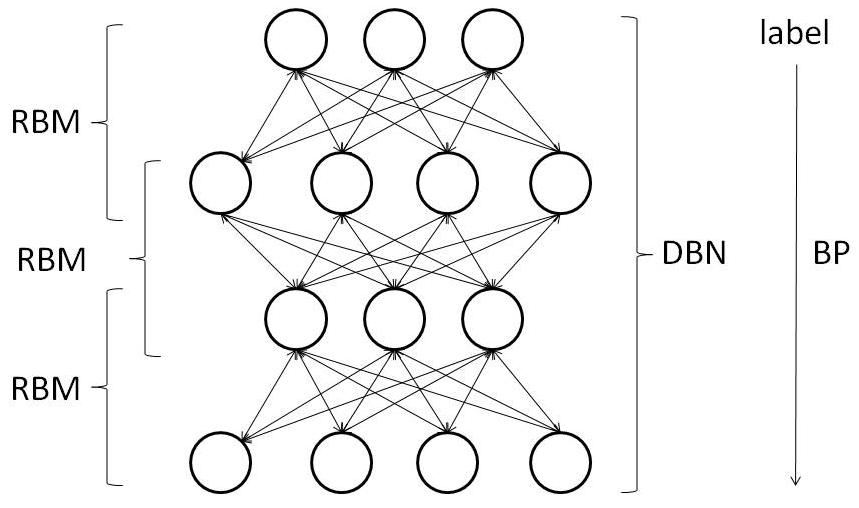}
\caption{\label{font-figure} The architecture of Deep Belief Networks.}
\end{center}
\end{figure}

In the experiment, we used DBN from Theano (Bergstra et al., 2010) and used Stochastic Gradient Descent (SGD) to optimize the loss function. We used held-out cross validation to tune the parameters. We fixed pretraining iteration to 25 for all word tasks. We determined finetuning iteration based on cross validation. We fixed pretraining rate to 0.1 and finetuning rate to 1. We got the best architecture of three hidden layers which have 100 hidden nodes for each layer. 

\subsection{Restricted Boltzmann Machine}
Restricted Boltzmann Machine (RBM) has one visible layer {\em v} and another hidden layer {\em h}. There are only edges with weights {\em W} connecting between nodes in different layers. This makes RBM a bipartite graph. 
\begin{figure}[h]
\begin{center}
\includegraphics[scale = 0.32]{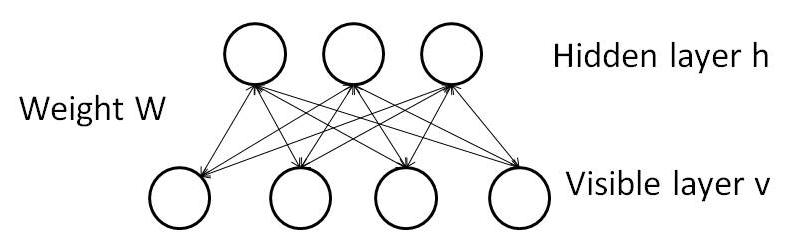}
\caption{\label{font-figure} The architecture of Restricted Boltzmann Machine.}
\end{center}
\end{figure}

Moreover, this also makes the hidden nodes to be independent given the visible node. 
\begin{eqnarray}
p(v|h) = \prod_ip(v_i|h) \\
p(h|v) = \prod_jp(h_j|v)
\end{eqnarray}
So, when data vector {\em v} is given, we can get an unbiased sample quickly from the posterior distribution. Thus, this eliminates explaining away effect in graphical models.  RBM tries to model {\em h} by reconstructing {\em v} with minimum error. RBM is modeled by an Energy Based function (EB). The energy function could be defined as
\begin{equation}
E(v,h) = -b_{v}'v - b_{h}'h - h'Wv,
\end{equation}
where \begin{math} b_{v} \end{math} and \begin{math} b_{h} \end{math} are biases for visible layer and hidden layer respectively. The activation function of RBM is as follows,
\begin{eqnarray}
P(h_i = 1 |v) = \frac{1}{1+e^{-b_{h_i}-W_iv}} \\
P(v_j = 1 |h) = \frac{1}{1+e^{-b_{v_j}-W_j'h}} 
\end{eqnarray}
Learning in RBM via maximum likelihood can be achieved by Gradient Descent but could be approximated by Gibbs sampling. RBM starts with taking an input vector as a visible layer. Then, RBM updates all hidden nodes simultaneously. After that, RBM tries to reconstruct the visible layer to get the reconstruction to update the hidden layer again.  The gradient term is as follows,
\begin{equation}
\Delta w_{ij} = \epsilon ({\langle v_i h_j \rangle}_{data} - {\langle v_i h_j \rangle}_{reconstructed}).
\end{equation}
\begin{figure}[h]
\begin{center}
\includegraphics[scale = 0.32]{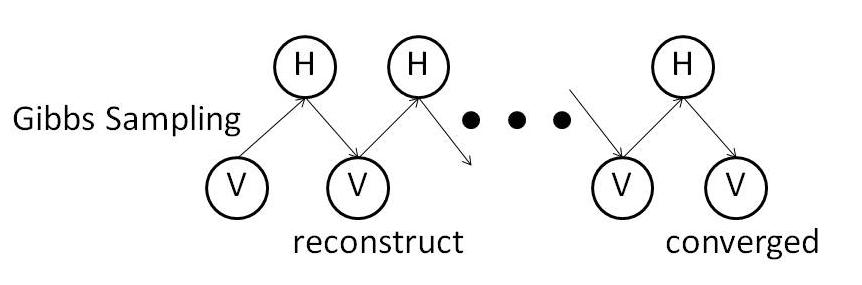}
\caption{\label{font-figure} Learning Restricted Boltzmann Machine.}
\end{center}
\end{figure}
This Gibbs sampling can be done iteratively until converge. However, this consumes a lot of computational power. So, Contrastive Divergence (CD) was proposed by (Hinton, 2002) to approximate this process by introducing KL-divergence such that performing Gibbs sampling only a few steps is enough. It was shown that only one Gibbs step is sufficient empirically. 

\subsection{Pretraining phase}
In WSD, features come from various knowledge sources and the classifier usually takes them without considering the relation among features. Hidden nodes of RBM are the combinations of features which provide a way to model those relations. By stacking RBMs, we can learn complex relations of knowledge sources. The learned hidden layer will be used as an input vector for another hidden layer. This phase is unsupervised and does not require labels.

\subsection{Finetuning phase}
After an unsupervised pretraining, the current parameter set is a good initial set to start local search by the backpropagation algorithm (BP). In pretraining, the search space is smoother and the optimal value is near the one in finetuning phase so this eliminates the possibility of stuck in poor local optimum. Backpropagation after pretraining works better because it works better in a large network since the gradient may become small but that slight changes of weights is enough to get a good model.

\section{Empirical Results}

\subsection{Data set and Evaluation}
In this paper, we used the SENSEVAL-2 data set which has 73 word tasks, 8,611 training instances and 4,328 test instances. All senses were labeled by WordNet 1.7. Our experiment is based on the official data set and fine-grained evaluation of SENSEVAL-2 which measured system performance by micro-average recall (mi). Moreover, we measured the significant by performing two sample one-sided t-test between DBN and other learning algotithms as in Table 2.
\begin{equation}
mi = \frac{number\, of\, correctly\, predicted\, instances}{number\, of\, all\, test\, instances}
\end{equation}
The baseline is Most Frequent Sense (MFS) which always chooses the major class of each word task as its prediction.
\subsection{Topical Feature}
In topical feature, one Nearest Neighbor algorithm (1-NN) performed lower than the baseline and dimensionality reduction techniques improved the performance only a little. The polynomial kernel tends to work better than the Gaussian RBF kernel in Kernel Principal Component Analysis (KPCA) but worked worse in Support Vector Machine (SVM) which only got a comparable performance as the baseline. Among shallow learning algorithms, Linear SVM worked best followed by Logistic Regression. Multilayer Perceptron (MLP), despite having hidden layer that made it able to model nonlinearity, worked worse than Logistic Regression. This goes with the argument of (Towell and Voorhees, 1998). In spite of small sample per class and noisy feature, Deep Belief Networks (DBN) achieved the best performance. DBN outperformed the baseline by 9.65\%, Logistic Regression by 2.07\%, MLP by 2.24\% and Linear SVM by 1.98\%.

% topical local pos all
\begin{table*}[t]
\begin{center}
\begin{tabular}{|l|r|r|r|r|}
\hline  \bf Learning Algorithm & \bf topical  & \bf local & \bf part-of-speech & \bf all feature \\ \hline 
MFS & 47.60\%(0.007) & 47.60\%(0.000) & 47.60\%(0.001) & 47.60\%(0.000) \\
1-NN & 38.08\%(0.000)  & 51.78\%(0.001) & 47.43\%(0.000) & 43.11\%(0.000) \\
PCA & 38.66\%(0.000) & 43.30\%(0.000) & 41.66\%(0.000) & 44.45\%(0.000)\\
KPCA(polynomial) & 38.22\%(0.000) & 32.12\%(0.000) & 45.82\%(0.000) & 37.50\%(0.000) \\
KPCA(Gaussian RBF)  & 35.95\%(0.000)  & 36.62\%(0.000) & 35.47\%(0.000) & 47.71\%(0.000) \\
NB & 50.16\%(0.028)  & 49.61\%(0.001) & 53.33\%(0.022) & 49.95\%(0.001) \\
Logistic Regression & 55.18\%(0.252)  & 57.26\%(0.198) & 54.86\%(0.039) & 60.07\%(0.267) \\
MLP & 55.01\%(0.251) & 57.26\%(0.129) & 54.83\%(0.043) & 59.70\%(0.224) \\
Linear SVM & 55.27\%(0.285) & 58.50\%(0.120) & 51.94\%(0.005) & 60.40\%(0.313) \\
SVM(polynomial) & 47.60\%(0.007) & 47.60\%(0.001) & 47.60\%(0.001) & 47.71\%(0.000) \\
SVM(Gaussian RBF) & 49.01\%(0.018)  & 48.43\%(0.000) & 47.92\%(0.001) & 51.02\%(0.003) \\
\bf DBN & \bf 57.25\%(-) &  \bf 61.23\%(-) &  \bf 60.07\%(-) & \bf 61.30\%(-) \\
\hline
\end{tabular}
\caption{\label{font-table} Micro-average recall (p-value compared to DBN (lower is better)) of various learning algorithms in  SENSEVAL-2 data set. }
\end{center}
\end{table*}

\subsection{Local Feature}
In local feature, 1-NN performed well by less noisy feature but dimensionality reduction techniques did not improve the performance. Gaussian RBF kernel tends to work better than polynomial kernel in KPCA and SVM. Naïve Bayes (NB) got worse performance compared to topical feature. The reason could be its strong independent assumption agrees with bag-of-words feature more than binary feature since bag-of-words feature also assumes independence among words in sentences. Among shallow learning algorithms, Linear SVM worked best followed by Logistic Regression and MLP. Moreover, they all got better performance compared to topical features since local feature is less noisy. DBN continued to achieve the best performance with better score than topical features by 3.98\%. In this feature, DBN outperform the baseline by 13.63\%, Logistic Regression and MLP by 3.97\% and Linear SVM by 2.73\%.

\subsection{Part-of-speech Feature}
In part-of-speech feature, 1-NN has lower performance than the baseline by a little. Dimensionality reduction techniques did not improve the performance. Gaussian RBF kernel tends to work better than polynomial kernel in KPCA and SVM. Na\"ive Bayes (NB) got the best performance compared to topical feature and local feature. Among shallow learning algorithms, Logistic Regression worked best followed by MLP and linear SVM. The performance was worse than local feature but better than topical feature. DBN continued to achieve the best performance with better score than topical feature by 2.32\% but worse score than local feature by 1.16\%. In this feature, DBN outperformed the baseline by 12.47\%, Logistic Regression by 5.21\%, MLP by 5.24\% and Linear SVM by 8.13\%. Compared to other features, DBN outperformed other shallow learning algorithms most significantly in this feature set.

\subsection{All Feature}
When these three features were combined, 1-NN performed worse because of noises and scarcity of the data. NB performed worse than local feature and part-of-speech feature but still better than topical feature alone. Logistic Regression and Linear SVM achieved the score higher than 60\%. This shows that adding features improved performance in both linear learning algorithms. However, MLP performed a little bit worse than Logistic Regression. In spite of small sample per class and sparse feature, Deep Belief Networks (DBN) still achieved the best performance of 61.30\%. DBN outperformed the baseline by 13.70\%, Logistic Regression by 1.23\%, MLP by 1.60\% and Linear SVM by 0.90\%. This did not have much improvement when compared to local feature alone. This may be concluded that using basic feature like local feature can make DBN achieve a fairly high performance without adding many features. 

\section{Summary and Future Work}
We have applied novel deep learning algorithm, namely, Deep Belief Networks (DBN) that makes an improvement to Word Sense Disambiguation (WSD) in term of accuracy. We evaluated three knowledge sources and compared with various state-of-the-art shallow learning algorithms whether they are linear or nonlinear. The experiment results show superiority of DBN over many state-of-the-art algorithms including Support Vector Machine (SVM). From Table 2, DBN outperformed the baseline, one Nearest Neighbor (1-NN), dimensionality reduction techniques (PCA and KPCA), Na\"ive Bayes (NB) and nonlinear SVM significantly. However, compared to Logistic Regression, Multilayer Perceptron (MLP) and Linear SVM, DBN significantly outperformed them in part-of-speech feature, slightly significant in local feature and not so much significant in topical and all features.

We also found that DBN achieved a relatively high performance when using only local feature while SVM needed more features. Thus, this indicates that deep learning algorithms help us extract useful properties from the data without excessive feature engineering. This shows that applying deep learning algorithms can be beneficial since there exists some nonlinearity in WSD data even if the data have a small number of instances and a lot of dimensions. The model of DBN that we got by cross validation shows that if the number of training instances is small, small architecture and large learning rate could be a good model.

This leads to future works in many directions. Firstly, sharing representation across word tasks can be helpful to improve the overall task since there is a few example per word task. Secondly, more knowledge sources including ones without label may be incorporated. We will investigate these further directions in the future.

\end{document}